# CEAI: CCM based Email Authorship Identification Model

Sarwat Nizamani[1,2], Nasrullah Memon[1]

[1]The Mærsck McKinne Møller Institute, University of Southern Denmark

[2]Department of Computer Science, Sindh University Campus Mirpurkhas, Pakistan

emal:{saniz,memon}@mmmi.sdu.dk

## Abstract

In this paper we present a model for email authorship identification (EAI) by employing a Cluster-based Classification (CCM) technique. Traditionally, stylometric features have been successfully employed in various authorship analysis tasks; we extend the traditional feature-set to include some more interesting and effective features for email authorship identification (e.g. the last punctuation mark used in an email, the tendency of an author to use capitalization at the start of an email, or the punctuation after a greeting or farewell). We also included Info Gain feature selection based content features. It is observed that the use of such features in the authorship identification process has a positive impact on the accuracy of the authorship identification task. We performed experiments to justify our arguments and compared the results with other base line models. Experimental results reveal that the proposed CCM-based email authorship identification model, along with the proposed feature set, outperforms the state-of-the-art support vector machine (SVM)-based models, as well as the models proposed by Iqbal et al. [1, 2]. The proposed model attains an accuracy rate of 94% for 10 authors, 89% for 25 authors, and 81% for 50 authors, respectively on Enron dataset, while 89.5% accuracy has been achieved on authors' constructed real email dataset. The results on Enron dataset have been achieved on quite a large number of authors as compared to the models proposed by Iqbal et al. [1, 2].

**Keywords** CCM; email authorship analysis; feature selection; machine learning; stylometric features; SVM

## 1 Introduction

Written communication that employs the use of computers is referred to as computer mediated communication (CMC), and email is one of the most important and widely used forms of CMC. Other types of CMC include online messages, blogs, forums and instant messaging services. Among all CMC, email has remained a key source of written communication, especially in the last few decades. Due to its salient features, it is the preferred source of written communication for almost every population connected to the Internet. It is a very quick, asynchronous written communication channel which is used for various purposes ranging from formal to informal communication. Formal emails include official correspondence, meeting calls, seminar calls, business communication, order placements, and flight reservations to name a few, while informal email communication includes personal emails, greetings and invitation emails (family and friends). These are all examples of the positive and legal applications of email; but, there also exist a number of illegitimate and negative purposes for email as well. As the Internet grows at a quicker rate each year, the same is also true for cybercrime [3]. According to Loader and Thomas[3], any illegal act using CMC is termed as a cybercrime. Email is a prime target for cybercriminals, as it has many unique features that enable the facilitation of criminal activity [4]. One can easily hide his/her identity while sending fraudulent emails, and anonymous messages can be sent to a number of recipients to spread fear among them. An email message can spread among a large population, across a large geographic location, literally in an instant. As we discussed above, email has been used for many illegal uses; the most prolific is the sending of "spam" mails [6], which are solicitous messages that the receiver is not interested in. Spam emails can be of various types, and usually spam emails are sent by unknown identities. Most of the spam we receive is sent for advertisement purposes. The other type of spam is sent specifically to obtain personal information (for example: the mail may outline a fraudulent contest for which the receivers may win a prize, if only they send the scammer their bank account information). Another use of spam email is "phishing" [7], or email that includes a link to a fraudulent website. These types of spam emails are sent with bad intentions, but even an average person may send a fake email for the sake of fun and then share it with others as well. There are certain service providers [1] that provide services for sending fake emails.

Apart from these spam emails, the 9/11 commission report [4] reveals that a number of emails were sent by the terrorists before the event took place. The investigations of the Mumbai attacks in 2008 [5] show that some emails were discovered revealing hard evidence of the attacks. This motivates us to put into place such a system that can reveal the identities of those criminals who intend to commence crimes using computer technology.

The study used to reveal the identities of illicit emails comes under the category of "authorship analysis." Authorship analysis [8, 9, 10] has been studied in different dimensions ever since it began, in 1887, when Mandenhal [11] analyzed the writings of Bacon, Marlowe and Shakespeare to distinguish their styles from one another. The authorship analysis study ranges from identifying the true author of a research article [12] to writer identification of literary texts [11]. It also ranges from revealing the identity of text messages that may be a post of a forum [9] or an email message [1, 2, 13]. Each type of writing has its unique characteristics, so the authorship analysis model designed for one kind of writing may fail to identify the author of another kind of writing. The reason is that some writings may be written by literary scholars, but others might be written by undereducated or near-illiterate authors. Therefore, there is a need for a research study for authorship identification separately, for each kind of writing.

In this paper we propose a CCM [14] -based EAI model (CEAI), which is essentially a cluster-based classification model for email authorship identification. First, emails are clustered, based on similar content , even though they may be discovered as written by different authors. For example, the use of similar words may be found in emails from different authors, but each author may have other different patterns, such as unique greeting styles or punctuations. Thus, in each cluster the number of authors is less than the original number of authors in the original email set. The classification technique, SVM [15] is then applied on each cluster. The classification results are improved when there are fewer classes; in our case, each distinguished author is a class. The classification then identifies the rest of the distinguishing patterns of different authors' emails in each cluster. This process is elaborated in Section 3 and shown in Fig 1.

We highlight the contributions of the paper as follows:
- ➢ Adaptation of a new text classification model (CCM) for email authorship identification
- ➢ The use of extended feature set for the EAI task
- ➢ The successful use of the feature selection technique for content features for the task
- ➢ Achieved high accuracy of the task for somewhat large number of authors

---

[1] http://www.sendanonymousemail.net/

In this paper, we have applied proposed model on a publicly available Enron email dataset[2] and on real emails dataset constructed by the authors, for experimental evaluations. A brief overview of Enron's email dataset is presented in the next Section, followed by the problem statement.

## 1.1 Datasets

The Enron email dataset [16] is a large collection of real emails of an organization, which once was a global organization. The original dataset contains 619,446 email messages belonging to 158 employees of the Enron organization. The dataset was put online by Willam Cohnen of CMU in March 2004. The dataset is processed by many researchers according to their requirements. The dataset used in this study is processed by Shetty and Adibi [17] and is available for download in a SQL dump[3] file at the University of Southern California's website[4]. Enron dataset is a benchmark dataset used for experimentation for various email related research. We also applied the proposed model on real email dataset (created by us) which is comprised of 7 authors and 30 emails of each author.

## 1.2 Problem Statement

We define our problem as assigning an author to an anonymous email after analyzing its features. Mathematically, the email authorship identification problem is represented as:

$$E = \{e_1, e_2, \ldots e_n\} \qquad (1)$$

$$A = \{a_1, a_2, \ldots, a_m\} \qquad (2)$$

$$F = \{f_1, f_2, \ldots, f_l\} \qquad (3)$$

$$EAI(F, E, A) = E_i \rightarrow A_j \qquad (4)$$

Where $E$ is a set of emails in the dataset, $A$ is the set of authors of the emails in the dataset, and $EAI(F,E,A)$ is a mapping function of email $e_i$ to author $a_j$, defined over $F$, $E$ and $A$, while $F$ is a set of features. Initially, during training, the emails of each author are analyzed and unique features are extracted while a training model is constructed. At testing time, the test email is checked against the features of all users. It is assigned an author who has greater similarity with the email being tested. The whole process is discussed in Section 3 and illustrated using algorithms.

The rest of the paper is organized as follows: The related work is discussed in Section 2, where the proposed model is described in Section 3. Experimental results and discussions are presented in Section 4 while the conclusion, along with future work, is given in Section 5.

## 2   Related Work

---

[2] http://www.cs.cmu.edu/~enron/

[3] My SQL dump file is a backup file of large database

[4] http://www.isi.edu/~adibi/Enron/Enron.htm

In this section we discuss related work in connection to CEAI. We also briefly elaborate the research dimensions in the email domain contained herein. The research in the email domain can be divided in two types; (1) the email traffic analysis [18]; and (2) the email content analysis [7, 19]. Email traffic analysis usually uses the email's header including the sender and receiver's email addresses, the email sending date, the type of receiver (To, Cc or Bcc) and the email subject. Email content analysis focuses on the email body and occasionally the subject field of the header. In this paper we deal with the content (body) of the email message, and the sender part of the email header. In the header, the research mainly focuses on Social Network Analysis [20] of an email network. It uses the links between the sender and the receiver of an email and sometimes uses the date of the link to conduct time stamping experiments. In the email content analysis, the research is mainly conducted using text mining and Natural Language Processing [21] techniques. Some of the dimensions for email content analysis are automatically categorizing emails [19] into user-defined folders, analyzing emails for spam [6]contents, analyzing emails for authorship identification [1], analyzing emails for phishing emails [7], etc.

Authorship analysis [10]can be characterized as regarding a section of writing based on its writing style to identify its true author. This analysis includes linguistics techniques, statistical analysis and machine learning techniques to draw a conclusion regarding the author of a particular writing. The authors [22] described some characteristics of authorship analysis, that are applicable to software forensics. Broadly, authorship analysis can be researched into three dimensions, namely; authorship identification, authorship characterization ,and similarity detection.

*Authorship identification* [10] can be described as identifying the author of a text from anonymous writing, based on the author's past writing records. Authorship identification has been previously researched as a text classification [23] problem in which the authors are each considered classes, and the writing is considered as text to be classified. It uses machine learning technique to decide whether a text has been written by a particular author or at least the likelihood that the text has been written by a particular author.

*Authorship characterization* [10] can be termed as creating author profile from an author's past writings. Sometimes this could include the author's gender, educational background and language familiarity as well.

*Similarity detection* [10] can be described as comparing some portion of a writing, without actually knowing the author and then deciding how many similarities are found in each writing style. This technique is mainly used as study in anti-plagiarism.

Authorship analysis research can be found early in its history, when Mandenhal [11] conducted a study to compare the writing styles of English authors Bacon, Marlowe and Shakespeare. Mosteller and Wallece [12] conducted experiments to identify the true authors for the disputed Federalist[5] articles that promoted the ratification of the U.S. Constitution .

The research conducted by de Vel et al. [24] considered the email authorship as a topic authorship categorization problem. A corpus of emails taken for the study was very small; only 156 emails were taken, all of them belonged to a newsgroup. Three authors and three topics were used in the study, as it was preliminary work in the email authorship area and showed promising results. The authors also proposed structural features that have since become a part of stylometric features for the task of authorship identification.

---

[5] http://en.wikipedia.org/wiki/Federalist_Papers

The authors [25] presented a study in which the importance of writeprints was evaluated for law enforcement to prevent cybercrimes. The authors claimed that the feature selection technique could use a cyber criminal's writeprint to identify them, much like using fingerprints as a form of identification in the "real" world.

The author [26] presented a detailed survey on authorship identification tasks. The study explored various aspects of the authorship task and research advances in the area. The survey first investigated the use of stylometric features for the task in various studies, then presented a detailed analysis of authorship attribution methods.

Koppel et al. [27] present detailed study on computational methods in authorship identification. The authors discuss authorship study literature in detail in connection to three main approaches used overtime, which include authorship characterization, similarity detection and authorship identification. Koppel et al. [28] present an approach for decomposing multi-author documents in authorial components. The technique used by the authors for the same purpose was an un-supervised machine learning technique.

A framework for authorship identification for online messages was presented by Zheng et al. [10], in which authors extracted four types of writing style features for e.g. lexical, syntactic, content specific and structural features and applied a machine learning classification algorithm for authorship identification. Authors considered authorship identification as a classification problem. High accuracy was achieved using a support vector machine [29] classifier; however, experiments were also conducted using a decision tree [30] and back propagation neural network [31].The authors conducted the series of experiments on English and Chinese online messages with a maximum of 600 messages used for experiments for both English and Chinese datasets. The study achieved promising results on the English dataset, and satisfactory results on the Chinese dataset. However, this dataset was very small.

In the literature, researchers attempted to explore new features overtime for the authorship identification task. With other writing style features, content specific features were also embedded for the authorship identification task by Iqbal et al. [1] and Zeng et al. [10]. The effectiveness of such features, like word usage frequency, was obvious, but it was realized that it would be topic specific. Still, these features are considered important for the authorship identification task.

If we look back to the earlier developments for authorship identification , Yule [32] identified that sentence length would be a useful feature for the authorship identification task. Yule [33], in his later work, he also suggested the richness of the vocabulary features to support the task. These features are still considered very useful for identifying a writing style.

The authors [8] applied authorship identification in a slightly different way, by applying stylometric authorship identification in the electronic market for scalability and robustness. The experimental results were successful by achieving good results.

The stylometric approach was applied by Abbasi and Chen [9] for both identity level identification and similarity detection, using the writeprint model for various online messages. In the study, the authors used an extended feature set in addition to baseline stylometric features, and performed a series of experiments on various online message datasets.

Iqbal et al. [13] developed a write-print based approach by mining frequent patterns. Write-print based approach considers a set of suspected authors for a malicious email. The frequent patterns are extracted from the malicious email and is matched against the frequent patterns of each suspected authors' emails and the most plausible author is assigned to that malicious email.

A cluster-based writeprint model for the email authorship was developed by Iqbal et al. [1] in order to evaluate on the Enron email dataset. The experiments were conducted using three clustering algorithms. The results given by the authors are on just a small portion of the Enron email dataset (500-1000 emails). The study [1] first clustered each author's emails, then extracted stylometric features and mined the frequent patterns from them. Afterwards, the frequent patterns were matched from the emails of different authors, and the study kept only those patterns in the model which were unique to each author. In the study, the authors reported the results on 5 and 10 authors only. The accuracy achieved was 90% for 5 authors and 80% for 10 authors respectively, where for each author, 100 messages were used. In another study of email authorship identification, Iqbal et al. [2] which is extension of Iqbal et al. [13] conducted experiments using up to 20 authors and the accuracy achieved was below 75%.

We propose an email authorship identification model using CCM, which is a cluster-based classification model. Better results are achieved on quite a large email dataset with just slightly higher number of authors. The model has been applied on 10, 25 and 50 authors and has achieved promising results. In this paper, no limit has been placed on the number of emails for each author because at investigation time it is not necessary that the number of emails available from each suspected author should be the same. The average number of emails per authors used for experiments is 600 and average number of words per email is 135. The proposed model CEAI achieved 94.3 % for 10 authors, 89% for 25 authors, and 81.3% for 50 authors, respectively. According to Iqbal et al. [2], the number of suspects under forensics investigation is usually considered less than 10, as the authors [2] mentioned in their discussions with a Law Enforcement Unit in Canada. Therefore, the results achieved using the CEAI model are 81.3 % for 50 authors -- quite a promising results.

Thus, here under we present some research questions that are answered by the proposed model; then we discuss the proposed model;

## 2.1 Research Questions

> ➢ Do extended features have a positive impact on EAI accuracy?
> ➢ Does Info Gain feature selection based content features have a positive impact on the accuracy of the model?
> ➢ Will CCM be able to attain high accuracy for the EAI task?
> ➢ Will the CCM-based EAI model be able to distinguish among many authors?

# 3   Proposed Model

This section is divided into subsections to outline the proposed model. An overview of the model is provided first, followed by a description of the feature set construction. Afterwards, the algorithmic details of the model are elaborated; finally the architecture of the model is illustrated, followed by some implementation details.

## 3.1 Overview of CEAI

The current study presents an Email Authorship Identification model based on CCM. The model uses a top-down approach to group the emails into clusters, although, the clusters created do not directly map to each author due to similarities in the writing styles of various authors. However, with the help of clusters, the authors who share a writing style are grouped. The classification algorithm SVM is then applied to each cluster to further separate each author's writing, as shown in Fig. 1. This way the authors are divided into clusters and in each cluster there are fewer authors. Thus, the classification model works well to classify the authors in clusters, and the overall accuracy of the EAI task increases by employing CCM on a slightly large number of authors as compared to the model [1, 2] on Enron dataset.

## 3.2 Feature Set Construction

As the proposed model uses machine learning techniques – clustering and classification – for the success of any machine learning model, it is worth to first construct a feature set. In the EAI task, features are used to represent the author as his/her writeprints. As fingerprints [1] have historically been used by law enforcement experts to uniquely identify criminals, writeprints [1, 9] could be used to identify authors to their writing styles. Researchers can apply their results, to cybercrimes, especially to fraudulent emails, by revealing the identity of the author. It was realized that, like unique fingerprints, each writer may have unique writing style. These unique writing style features are termed as "stylometric" [9] and have previously been used in studies [1, 9, 10, 34], to identify authors using lexical, structural and syntactic features. In Table 1, we provide a list of some stylometric features that are commonly used in an authorship identification study.

Table 1: Stylometric features

| Lexical Features | Description |
| --- | --- |
| Email length in characters | Average number of characters in emails by an author |
| Digit density | The ratio of digits to total no. of characters in email an author's emails |
| Ratio of space to email length | The ratio of space character to total length |
| Each character count | Normalized frequencies of each character |
| Each special character count | Normalized frequencies of each special character |
| Document length in words | Average number of words per email by an author |
| Average length of words | Average word length used by an author |
| Average sentence length | Average size of sentence in email of an author |
| Ratio of short words | Ratio of short words i.e. 3 character and less to total number of email words |
| **Structural Features** | |
| Number of paragraphs in documents | Average number of paragraphs in the emails of an author |
| Number of sentences in paragraphs | Average number of sentences in the emails of an author |
| Indentation of paragraphs | How paragraphs are usually indented by the author (tabs/space count) |

| Use of greetings in beginning | Use of specific greetings in beginning of the emails of an author |
|---|---|
| **Syntactic features** | |
| Frequencies of function words | Normalized frequencies of each function word in the emails of an authors |
| Frequencies of punctuations | Normalized frequencies of punctuation used by an author |

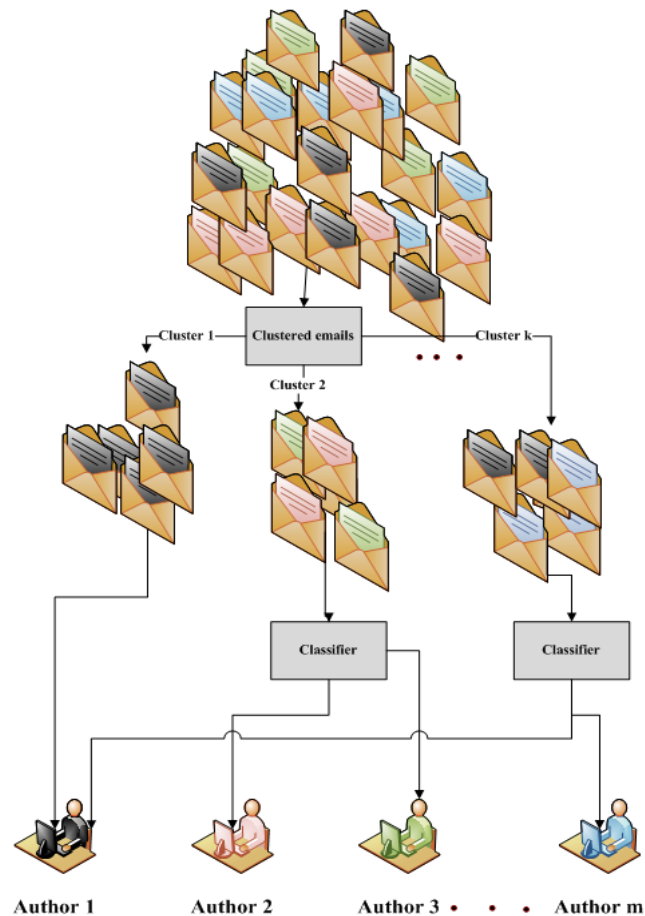

Fig. 1 CEAI process

The stylometric features are the unique features found into the writing style of an individual. These features reveal the writing patterns of an individual. The stylometric features have been divided into three main categories, namely; lexical, structural and syntactic features

Other features used for authorship identification are content specific. Content specific features vary from text to text and are extracted from the text of the email's body. In CEAI, for content specific features, we assigned weights to each term in the email's body. Weights have been assigned using the *tf-idf* (term frequency-inverse document frequency) transformation scheme [35]. The *tf-idf* scheme assigns high weights to each term in an email that is rare in many authors; this means that these terms have more discerning factor in distinguishing the emails of various authors.

The *tf-idf* works as follows

$$tf(t,e) = \frac{f(t,e)}{|e|} \qquad (5)$$

$$idf(t,e) = \log \frac{|E|}{1 + \{e : t \in e\}} \qquad (6)$$

$$tfidf(t,e) = tf(t,e) * idf(t,e) \qquad (7)$$

Where *f(t,e)* returns the frequency of term *t* in email *e* and */e/* is the size of the email body in terms of number of words; therefore, *tf(t,e)* returns the normalized frequency (in the range of 0 and 1) of term *t* in email *e*. The function *idf(t,e)* returns the inverse document frequency. This assigns highest weight to the terms in emails which are present in emails of few authors. */E/* is the total number of emails in the dataset, while *{e: t ∈ e}* returns the number of emails that contain term *t* in the body. In the content-specific feature set, we keep only the terms that have the minimum frequency 3. Still the number of these terms is too high, so we applied the feature selection using Info Gain[36]. Equations (8)-(10) give mathematical representation of the Info Gain feature selection method, which assigns ranks to the content terms. We selected the top-ranking 1000 terms by reducing up to 90% of the content features and maintaining the accuracy of the model.

$$IG(A, f) = H(A) - H(A \mid f) \qquad (8)$$

$$H(A) = -\sum_{i=1}^{m} p(A = i) \log_2 P(A = i) \qquad (9)$$

$$H(A \mid f) = -\sum_{i=1}^{m} p(A = i \mid f) \log_2 P(A = i \mid f) \qquad (10)$$

Where *H(A)* is the entropy of author *A*, *m* is the total number of authors in the dataset. The *IG(A,f)* is an Info Gain function that returns the importance of feature *f* as a distinguishing feature for author *A*. Besides the use of the above given features, our manual analysis yields the use of the few additional features that increase the effectiveness and accuracy of the proposed model.

- Frequently used farewell words before signature
- Ending punctuation
- Frequently used punctuation
- Modern features: Sent from mobile device feature
- Use of repetitive words (me,me,me/ yes,yes,yes)
- Time format 2:00, 2 o' clock
- Single sentence / questioning email
- Punctuation after and before farewell
- Punctuation after greeting
- Punctuation in short or incomplete sentences

We show the effectiveness of these features in comparison to baseline features in experimental results in Section 4.

*3.2.1    Implementation details of features*

For lexical and structural features extraction Java feature functions are created, where for syntactic features commonly used function words and punctuations have been used that have the capability of distinguishing the authors. We extracted the content specific features using WEKA [37] which is a freely available, open source machine learning tool.

We implemented Java classes for mapping each feature type to the corresponding Attribute Relation File Format (ARFF) which is a WEKA-specific data file. Lexical features (character based/ word based) were extracted as the ratio of the frequencies to the total number of characters/words. In other features, such as the last punctuation feature, five indexes are reserved. The index of the full stop is turned on when the email contained full stop at the end of the email body while all the other indexes of the last punctuation feature are turned off. In continuation to the implementation details, we discuss algorithmic details of the model in the next Section.

## 3.3 Algorithmic Details

In this section some algorithmic details of the proposed model are provided. The DataCleaning algorithm (Algorithm 1) is designed to extract the required part of the dataset, which is necessary for experimentation. The algorithm cleans the dataset and filters out the meaningless parts from the original dataset. In an un-cleaned dataset there are many pieces of information which are not needed for the authorship identification task. For experiments, we retrieved the body (Step 6 of the algorithm) of the email and from the header, the sender name is retrieved (Steps 3 and 4); in other words the author of email. The forwarded emails (i.e. the emails that did not belong to the sender) and also the referenced text from the emails has been removed (Steps 7 to 14). The author signatures have also been removed (Steps 15 and 16) because an anonymous author would never use his or her true identity. After cleaning the dataset, we applied some preprocessing steps necessary for the authorship identification task. These steps included tokenization, filtering greetings and farewell tokens. The stop words (function words) are not removed because the habitual use of these words provided clues for the identity of the true author of the email. In the algorithm the functions *GetHeader( ),GetSender(), GetReftype(), GetBody()* anf *GetSginature,* were applied using SQL queries to retrieve the header, sender, reference type, body and signature of the email message, respectively. The *RemoveReftext( )* is the function applied to remove the referenced part of the message that is not part of the current message body; and the function *Delete( )* is used to simply delete the forwarded email. Both of these functions are applied using SQL queries.

When the emails of different authors were manually analyzed, it was realized that the ending punctuation mark feature, in combination with other features, could be helpful for identifying the author of an email. Algorithm 2 describes the steps for extracting the last punctuation mark of the email body. In the algorithm, Steps 4 through 14, check and extract the various punctuation marks at the end of the email body.

In addition to our extended features, the baseline features used in the study are similar to Abbasi and Chen [9]. It was observed that these unique features, found in the emails written by one author but not found in the emails written by other authors have positive impact on the authorship identification task. The extended features included stylistic as well as content-based features. After cleaning and preprocessing, we extracted the features, then applied the feature selection and applied CCM for the EAI task. Algorithm 3 has been designed to extract different feature types. Initially the feature set is empty, then various feature types are extracted by separate functions such as the *Construct_Lexical_features( ),Construct_Structural_features()*

*Construct_Syntactic_features()*, which are used to extract lexical, structural and syntactic features respectively. *Extract_Content_features()* extracted content-based features. Finally, all the features are added to the feature set and saved to an ARFF file format. Among other features listed in Section 3.2, we present algorithm 2 which extracts the last punctuation type feature.

Cluster creation is the main part of the CEAI model; hence *BuildCluster* (Algorithm 4) creates k-clusters of emails using k-means [38] clustering algorithm. The parameter k is selected using visualization, which is one of the most common method [39] for selecting value of k because it is considered to be useful for validating clustering results [40]. The function *sim(c[i], c[j])* returns the similarity score between clusters, and it returns high score when emails in both clusters share common authors. At the time clusters are created, authors of emails are not kept in account; the algorithm groups the emails into clusters sharing various features. When merging is performed, the authors of the message are kept into account, and those clusters are merged that contain emails from common authors. It may be noted that when there is no optimized cluster, then the clusters are merged. An optimized cluster is one, whose size is greater than the threshold (the number of samples emails is at least twice of the number of authors in the cluster) and the cluster contains emails from subset of authors. For a better understanding of the model, interested readers may refer to the authors' already published work [14].

In order to apply the CCM model using various features, the final algorithm EAI (Algorithm 5) is developed. The algorithm receives three parameters as input E, F and A. The algorithm discusses all the steps of the *EAI* task and it is the main algorithm of CEAI model. The function *BuildCluster* creates clusters (Step 1) and the parameter k is determined by using visualization. The Steps 2 through 27 are run for each cluster, first the test is conducted if the cluster is comprised of all the emails belonging to single author, if so, the emails are assigned that author and cluster is returned in the final model (Steps 3-7 ). Next a test is conducted if the cluster size is less than a threshold, if so, the cluster is merged in most similar cluster (Steps 8-18). When a cluster is determined, whose size is greater than the threshold and not all emails belong to one author, a classifier is trained for that cluster and if its accuracy is higher than its parent classifier, the classifier is added in the final model and the classifier is considered to be the optimized classifier. If classifier does not return optimal results, further step of clustering takes place and the process is repeated.

The authorship identification task usually is concerned with a provision of samples of candidate authors' sample writing and the task is to determine the most plausible author of the test sample among the candidate authors. It should be noted that in this paper the proposed model is limited to the author identification task. In order to address the problem of unknown author, a non-parametric statistical test [41] is to be adapted. The test is referred as Kruskal-Wallis test [42], which is a non-parametric method to rank each feature with respect to all authors in the training set. Once an email has been assigned an author using the proposed model, the test is applied using similarity metric, to check whether the email belongs to the author's word distribution. If the email's word distribution differs from the assigned author's emails in the training set, the email is recognized to belong to an unknown author. The test is applied only when the training emails from the assigned author are used. In the Kruskal-Wallis test, the null hypothesis is rejected which articulates that for all the authors, the words' distribution remains same [42].

| Algorithm 1.  DataCleaning (Dataset) |
| --- |
| 1:  Cleaned_Dataset= null |

| | |
|---|---|
| 2: | *for* (i = 1 *to* n) |
| 3: | Header=   GetHeader (email[i]) |
| 4: | Sender=GetSender(Header) |
| 5: | Rtype=GetReftype(Header) |
| 6: | Body= GetBody(email[i]) |
| 7: | *if*( rtype = Re) |
| 8: | Body = RemoveReftext(Body) |
| 9: | *else if* (rtype = Fwd) |
| 10 | *and* ( (append Body = FYI ) |
| 11 | *or* (append Body =      NULL)) |
| 12: | *Delete* (email[i]) |
| 13 | ***Continue*** |
| 14: | *end if* |
| 15: | Signature=GetSignature(Body) |
| 16: | Body=Body-Signature |
| 17: | email[i]=Sender+Body |
| 18: | Cleaned_Dataset= Cleaned_Dataset + email[i] |
| 19: | *end for* |
| 20: | *End* |

**Algorithm 2. LastPunctuation(body)**

| | |
|---|---|
| 1: | intialize full_stop=null, q_mark=null, |
| 2: | exclam_symb=null, semi_coll=null, |
| 3: | comma=null |
| 4: | *if* (last_punct = '.') |
| 5: | full_stop=1 |
| 6: | *else if*(last_punct = '?') |
| 7: | q_mark = 1 |
| 8: | *else if* (last_punct = ',') |
| 9: | comma=1 |
| 10: | *else if*(last_punct = '!') |
| 11: | exclam_symb=1 |
| 12: | *else if*(last_punct=';') |
| 13: | semi_coll=1 |
| 14: | *end_if* |
| 15: | *End* |

**Algorithm 3. FeatureFunctionConstructor(Dataset)**

| | |
|---|---|
| 1: | feature_set=NULL |
| 2: | Lexical_features= Construct_Lexical features() |
| 3: | Structural_features= Construct_Structural_ features() |
| 4: | Syntactic_features=Construct_Syntactic_features() |
| 5: | Content_features= Extract_Content_features |
| 6: | feature_set=Lextical_features+Structural_features+... |
| 7 | Syntactic_features+Content_features |
| 8: | writeArff: features_set |
| 9: | *End* |

**Algorithm 4. BuildCluster(E,K)**

| | |
|---|---|
| 1: | C[1..k]=K-means(E,k) |
| 2: | *if*(sim (C[i], C[j]) > th) |
| 3: | Merge (C[i], C[j]) |
| 4: | *end if* |
| 5: | *End* |

| Algorithm 5. EAI(E,A,F) | |
|---|---|
| 1: | BuildClusters(E, k) |
| 2: | *for each* cluster c(i=1 *to* k) |
| 3: | *if* ($c_i$ contains emails that belong only to one author a) |
| 4: | *for each* email e in $c_i$ |
| 5: | author(e)=a |
| 6: | *end foreach* |
| 7: | *add*($c_i$,CEAI) //add cluster center to CEAI |
| 8: | *else if* ($|c_i| < \mu$)    // $\mu$ is threshold |
| 9: | max= 0.0 |
| 10: | *for each* pairs of clusters c (j=2 to k) //all clusters except $c_i$ |
| 11: | *if* (max < similarity($c_i$, $c_j$)) |
| 12: | max = similarity($c_i$, $c_i$) |
| 13: | cl=cj |
| 14: | m=j    //current most similar cluster |
| 15: | *end if* |
| 16: | *end foreach* |
| 17: | $c_i$=*union*(cl, $c_i$) |
| 18: | *delete*($c_m$) |
| 19: | *else* BuildClassifiers($c_i$, SVM) |
| 20: | *if* (isOptimized(SVM($c_i$)) |
| 21: | *add*( SVM($c_i$),CEAI) |
| 22: | *else* |
| 23: | BuildCluster(k,$c_i$) |
| 24: | *goto* 2 |
| 25: | *end if* |
| 26 | *end if* |
| 27: | *end foreach* |
| 28: | *End* |

## 3.4 CEAI Architecture

The architectural view of the proposed model describes the systematic processing of the model by illustrating the interaction of various modules. The architecture of CEAI comprises of input, preprocessing, vector input and an EAI module (CCM model) as shown in Fig. 2. The input module receives raw emails in a SQL dump file (compressed database file), the dump file is then transformed to SQL tables. The Input module then retrieves emails from the tables and makes them available in a form that could be processed by the preprocessing module. The emails received from the input module contain unnecessary data which is not required for email authorship identification process. It is the responsibility of data cleaning module to remove needless data and make available the required email data. As an email is comprised of two main parts; the header and the body, first the data is cleaned by retrieving the header and the body of the email. Afterwards, the sender is retrieved from the header, depending on the reference type, the body is cleaned. The body is cleaned by removing forwarded information if any forwarded piece of data is in the email; replied part of the email is removed if an email is reply of another email in order to preserve only the sender's text.

Once cleaned data is available, features regarding the EAI task are extracted by the feature extractor sub-module. The lexical, structural and syntactic features are extracted using Java code, while the content features are extracted using WEKA preprocessing module. Once all feature types are extracted in preprocessing module, all the feature types are then combined and converted to a feature vector. The feature vector is then given as an input to authorship identification module which uses CCM based approach for email authorship identification.

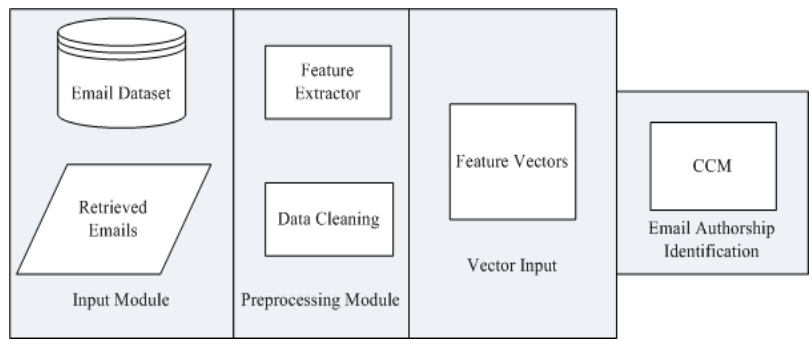

Fig. 2: CEAI Architecture

## 3.5 Implementation Details

The architectural design section described the basic modules of CEAI. In this section, we discuss the implementation details of each of the module. There are three types of implementations required for the proposed model; (a) dataset retrieval (input) and data cleaning which is performed using MySQL server; (b) feature extraction is the second type of implementation, which is implemented using Java code and ARFF files are generated which are then given to WEKA for applying the proposed approach; (c) the third type of implementation is to apply the proposed approach using WEKA, by first clustering and then classification.

Initially the required part of the dataset is extracted using the MySQL queries. For extracting the sender's text from the email and removing forwarded text and referenced text, MySQL queries are used. Once unnecessary parts of the dataset are removed, feature extraction is performed. Content-specific features are extracted using WEKA's StringToWordVector filter by assigning weights to each content term, then the Info Gain features selection is applied to rank the content terms according to author identification capability. For rest of the feature extraction, Java code has been implemented, which also combines all of the features in ARFF format, which is comprised of feature vector for each email. When dataset is available as feature vectors, k-means [38] clustering algorithm has been applied using WEKA on the ARFF file and on each optimized cluster, the SVM [15] algorithm is run from within WEKA. After presenting details of the proposed model, the experimental results and evaluation are discussed in the next Section.

## 4   Experimental Results and Evaluation

This section is divided in to sub-sections : Section 4.1 describes the experimental setup; Section 4.2 discusses the experimental evaluation, and Section 4.3 includes the results and a discussion.

## 4.1 Experimental Setup

In order to perform experiments, we retrieved just the "Sent" emails of the Enron employees from the dataset, because the true identities of the authors are available with their names. First, we randomly selected 50 authors, then retrieved the "Sent" emails of those 50 authors. We retrieved the email bodies from the message table and along with the sender names by joining queries on the Message and the Employee_list tables from the database. We performed five types of experiments on the emails of 50 authors for the authorship identification task. The first experiment is comprised of baseline stylometric features and an SVM classifier. The second set of experiments is conducted using baseline stylometric features with extended features and an SVM classifier for the author identification. In the third set of experiments we used baseline stylometric features with the Info Gain feature selection based content features. In the fourth set of experiments we used extended baseline stylometric features with Info Gain feature selection based content features. In the last set of experiments, we used extended baseline stylometric features, with Info Gain feature selection based content features, and for authorship identification we used CCM. We then retrieved the emails of 25, and then 10 authors respectively, and also performed all five types of experiments on both 25 and 10 authors. For authors' constructed dataset all of the five kinds of experiments have been performed.

## 4.2 Evaluation Criteria

The evaluation criteria used in our experiments is stratified 10 fold cross validation, as the same is used in study [2]. The 10 fold cross validation is a method by which the dataset is divided into ten subsets. Algorithms run for ten rounds; in each round, one subset is used for testing while the other nine are used for training. In each round, a different subset is taken for testing. Finally, the average of accuracy rate is calculated to find the final accuracy rate. This method of evaluation is considered to be more effective because each email is considered for its author to be predicted, thus every email is used exactly once in test set and nine times in the training set for each set of experiments. The final accuracy of all the runs is returned as average.

### 4.2.1 Evaluation measures

The evaluation measure used in our study is the accuracy, which is calculated as:

$$Accuracy = \frac{TP + TN}{TP + TN + FP + FN} * 100 \qquad (11)$$

Where $TP$ is the number of authors correctly classified to be the positive classes, whereas $TN$ is the number of authors correctly classified to be the negative classes, while FP is the number of authors incorrectly returned by the model as positive classes, and $FN$ is the number of authors incorrectly returned by the model as negative classes, where in fact they were positive. In general, TP+TN is the number of emails with correctly identified authors and TP+TN+FP+FN is the total number of emails. The accuracy of CCM for EAI is measured as the average rate of accuracy which is the percentage of correctly identified authors of emails . It is given by the following equation:

$$Accuracy_{CCM} = \frac{1}{n} \sum_{i=1}^{k} (|c_i| * ac_i) \qquad (12)$$

Where $k$ is the number of the clusters in the model; $|c_i|$ is the number of emails in cluster $i$; $ac_i$ is the accuracy of classifier in cluster $i$; and $n$ is the total number of emails.

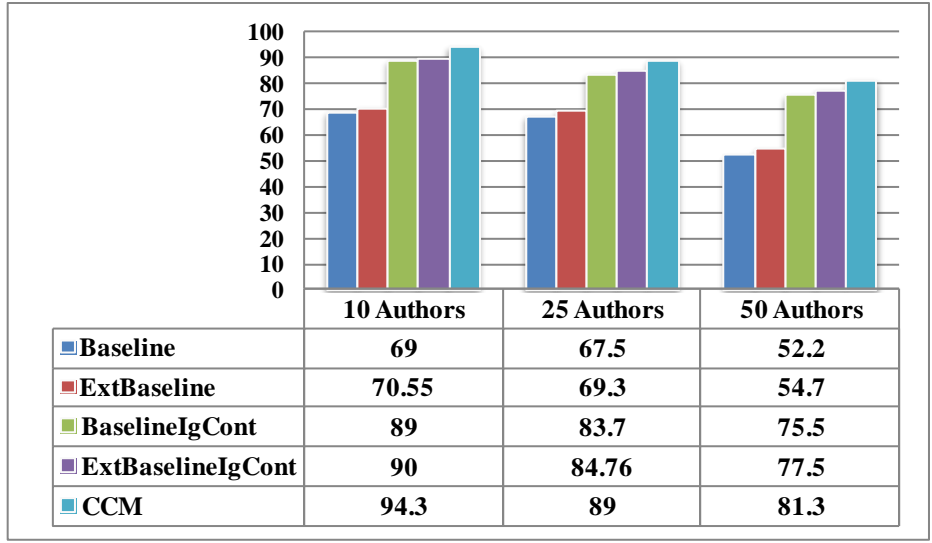

Fig 3. Experimental Results of CCM based EAI for 10, 25 and 50 authors

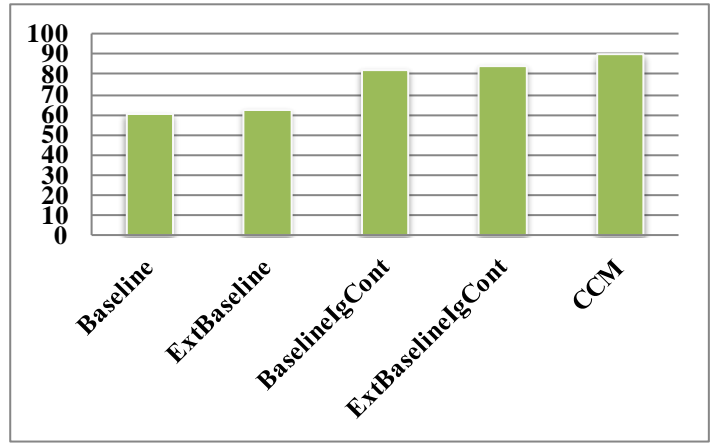

Fig 4. Experimental results of CCM based EAI on authors' dataset for 7 authors

## 4.3 Results and Discussion

The experimental results for 10, 25 and 50 authors on benchmark Enron dataset are given in Fig. 3. The results clearly depict that the model CEAI achieves the highest accuracy for the email authorship identification task on 10, 25 and 50 authors. We present the summary of all experiments' types in Table 2.

In each experiment, various feature sets are incorporated step-by-step and starting with the baseline stylometric features, which are traditionally used for an authorship analysis task. Afterwards, extended features, which slightly improved the accuracy of the task, are used. Info Gain based content features are then added, which keeps only the content features which have the capability of distinguishing each author. The original number of content features is too large and the Info Gain based content feature selection maintains the balance between dimensionality reduction and accuracy of the task. The accuracy of the task increased greatly with the addition of these features. The first four experiments are performed using SVM and the last experiment has been conducted using CCM on the final set of features that again improves the accuracy of task. All five types of experiments have been performed on authors' constructed real email dataset of seven authors. Analogous effects of all types of experiments have been observed on this dataset; though a bit less accuracy has been achieved, due to the fact that for this new dataset we have smaller amount of emails for each author as compared to the Enron dataset.

Table 2. Summary experiments

| Exp. No. | Feature Set | Model |
|---|---|---|
| 1 | Baseline | SVM |
| 2 | Extended Baseline | SVM |
| 3 | Baseline plus IG feature selection based content features | SVM |
| 4 | Extended Baseline plus IG feature selection based content features | SVM |
| 5 | Extended Baseline plus IG feature selection based content features | CCM |

We achieved highest accuracy, 94%, on 10 authors, whereas accuracy decreased to some extent for 25 authors that is 89%, and 50 authors which is 81%, respectively on Enron dataset; while 89.5 % accuracy has been achieved on authors' constructed dataset.

## 5   Conclusion and Future Work

In this paper we present a model for email authorship identification (EAI) and its evaluation on the publicly available Enron email dataset as well as authors' constructed dataset. The contributions of the paper are twofold: (i) to evaluate the use of additional features with baseline stylometric features for the EAI task and Info Gain feature selection based content features; and (ii) the development of a new model for the EAI task. Our experiments show that additional features have positive impact on authorship identification accuracy, and the evaluations confirm that CEAI achieves an accuracy of 94% on 10 authors, 89% on 25 authors, and 81% on 50 authors and 89.5% accuracy has been achieved on authors' dataset of seven authors. The promising results are achieved as compared to previous models [1, 2] on benchmark Enron dataset and used a slightly larger number of authors in the experiments (50 authors as compared to 20 authors). It has been observed that the accuracy of the task decreases to some extent with the increase of authors. It is still necessary to improve the email authorship identification task by introducing new features or by improving the model so that it may be scalable to an unlimited number of email authors. At present the proposed model is capable for authorship identification, but in future we would like to extend the model to authorship verification in case if an email belongs to anonymous author, to identify the email as unknown.   We plan to extend our experiments of the authorship identification on other online social media, especially in forums in English as well as in Urdu languages.